\documentclass[journal]{IEEEtran}
\usepackage{latexsym}

\usepackage{enumerate}
\usepackage{graphicx}
\usepackage{subfigure}
\usepackage{multirow}

\usepackage{framed,multirow}
\usepackage{times}
\usepackage{graphicx} 
\usepackage{subfigure}
\usepackage{amssymb}
\usepackage{latexsym}

\usepackage{algorithm}
\usepackage{algorithmic}

\usepackage{hyperref}

\usepackage{url}
\usepackage{xcolor}
\usepackage{amsmath}
\usepackage{amsxtra}
\usepackage{braket}
\usepackage{array}
\usepackage{color}
\usepackage{url}
\usepackage{placeins}

\usepackage{times}
\usepackage{balance}
\usepackage{enumerate}
\usepackage{multirow}
\usepackage{pifont}
\usepackage{marvosym}
\usepackage{ifsym}
\usepackage{threeparttable}

\usepackage{lipsum} 
\usepackage{amsmath,esint} 
\usepackage{cuted}

\begin{document}
\title{ENADPool: The Edge-Node Attention-based Differentiable Pooling for Graph Neural Networks}
\author{Zhehan~Zhao,~\IEEEmembership{}
        Lu~Bai${}^{*}$,~\IEEEmembership{IEEE~Member}
        Lixin~Cui,~\IEEEmembership{IEEE~Member}
        Ming~Li,~\IEEEmembership{IEEE~Member}\\
        Yue~Wang,~\IEEEmembership{}
        Lixiang~Xu,~\IEEEmembership{}     
        Edwin R.~Hancock,~\IEEEmembership{IEEE~Fellow}

\thanks{

Zhehanzhao and Lu Bai (${}^{*}$Corresponding Author: bailu@bnu.edu.cn) are with School of Artificial Intelligence, Beijing Normal University, Beijing, China. Lixin Cui and Yue Wang are with School of Information, Central University of Finance and Economics, Beijing, China.Ming Li is with the Key Laboratory of Intelligent Education Technology and Application of Zhejiang Province, Zhejiang Normal University, Jinhua, China. Edwin R. Hancock is with Department of Computer Science, University of York, York, UK}
}
\markboth{}%
{Shell \MakeLowercase{\textit{et al.}}: Bare Demo of IEEEtran.cls for Journals}

\maketitle

\begin{abstract}
  Graph Neural Networks (GNNs) are powerful tools for graph classification. One important operation for GNNs is the downsampling or pooling that can learn effective embeddings from the node representations. In this paper, we propose a new hierarchical pooling operation, namely the Edge-Node Attention-based Differentiable Pooling (ENADPool), for GNNs to learn effective graph representations. Unlike the classical hierarchical pooling operation that is based on the unclear node assignment and simply computes the averaged feature over the nodes of each cluster, the proposed ENADPool not only employs a hard clustering strategy to assign each node into an unique cluster, but also compress the node features as well as their edge connectivity strengths into the resulting hierarchical structure based on the attention mechanism after each pooling step. As a result, the proposed ENADPool simultaneously identifies the importance of different nodes within each separated cluster and edges between corresponding clusters, that significantly addresses the shortcomings of the uniform edge-node based structure information aggregation arising in the classical hierarchical pooling operation. Moreover, to mitigate the over-smoothing problem arising in existing GNNs, we propose a Multi-distance GNN (MD-GNN) model associated with the proposed ENADPool operation, allowing the nodes to actively and directly receive the feature information from neighbors at different random walk steps. Experiments demonstrate the effectiveness of the MD-GNN associated with the proposed ENADPool.
\end{abstract}


\maketitle
\IEEEpeerreviewmaketitle

\section{Introduction}

Convolutional Neural Networks (CNNs) are powerful tools to deal with the data that has a grid-like structure, e.g., the images and time series. However, the real-world data in many fields usually preserves the complicated structure information between components and cannot always be represented in regular formats, e.g., the social networks, the molecular structures, etc. Such irregular data can be typically represented as graphs. To effectively deal with the graph data, the CNNs have been further generalized for the irregular graph data. These methods are the so-called Graph Neural Networks (GNNs)~\cite{xu2018powerful} and have been widely employed for various real-world applications~\cite{wu2022graph,li2017diffusion,sun2020graph}.

Generally speaking, the GNNs can encode the high-dimensional sparse graph-structured data into the compact low-dimensional vector, simultaneously preserving the global graph structure characteristics and the local node information. Most of the GNN models~\cite{xu2018powerful,kipf2016semi,hamilton2017inductive,velivckovic2017graph} adopt the neighborhood aggregation strategy to obtain the structure representations. This strategy can be summarized as the following two steps, i.e., (1) Initialization: the node representations are initialized using their attributes or predefined rules, and (2) Updating~\cite{xu2018powerful}: the node representations are iteratively updated by aggregating the representations of their neighboring nodes. After the $k$ iterations of the aggregation, the node representation can effectively capture the structure information within its $k$-hop neighborhoods, and the resulting GNN models can provide powerful tools for graph data analysis, such as the node classification~\cite{grover2016node2vec}, the link prediction~\cite{zhang2018link}, and the graph classification~\cite{lee2019self}.

One challenging arising in the GNNs is that the GNN models need to abstract features from original graphs with less redundant information, when facing the graph-level tasks. To this end, the graph pooling operation becomes very crucial and can generate an overall representation of the graph structure. The main purpose of the graph pooling is to downsize the graph structure and make the representation learning effective. Most existing graph pooling methods can be categorized into two categories. The methods belonging to the first category are the global pooling methods, that tend to directly sum up or average all the node representations over the whole graph. Since the global pooling methods neglect the feature distribution associated with different nodes, they cannot identify the structural differences between the nodes. To overcome the shortcoming, the methods falling into the second category, namely the hierarchical pooling methods, are developed to capture the hierarchical structures through the features of local nodes by progressively reducing the graph size. Specifically, the reduction process of these hierarchical pooling methods can be achieved through two strategies, i.e., the TopK-based strategy~\cite{gao2019graph} and the cluster-based strategy~\cite{ying2018hierarchical,wu2022structural,bianchi2020spectral}. The TopK-based strategy employs a scoring function to evaluate the importance of different nodes in the original graph. Based on the importance scores, one can select the top $K$ nodes as the nodes of the resulting coarsened graph. On the other hand, the cluster-based strategy assigns nodes into different clusters using a learnable assignment matrix, and merge the nodes within each cluster as the compressed nodes of the resulting coarsened graph. Since the TopK-based methods may result in information loss due to the discarding of nodes \cite{ju2024comprehensive}, the cluster-based methods are usually more effective.

Unfortunately, most cluster-based hierarchical pooling methods are based on the soft node assignment, i.e., one node can be assigned into different clusters associated with different probabilities. Thus, the process of the node clustering may disrupt the completeness of the original node features. Moreover, these pooling methods tend to adopt an uniform aggregation approach when compress the node features and edge connectivity strengths, disregarding the importance of individual elements. To overcome these problems, in recent years the self-attention mechanism has been used~\cite{vaswani2017attention}. For instance, the ABDPool~\cite{liu2022abdpool} has adopted the self-attention mechanism to adaptively determine the importance of different nodes within the same cluster, and the nodes are assigned into a cluster based on a hard assignment matrix, i.e., each node can only be assigned into one unique cluster. However, the ABDPool involves traversing each cluster and then applying the dot product attention to the nodes of each individual cluster, making the computation expensive. Furthermore, the ABDPool only focuses on the node-based attention and cannot identify the importance of different edges between node clusters, that will be merged as the compressed edges of the resulting coarsened graph. In a summary, developing effective hierarchical pooling for GNNs still remains challenging problems. 

Another challenging arising in the GNN models is the notorious over-smoothing problem. This shortcoming is caused by the neighborhood aggregation strategy of the GNN models, and the nodes can continuously propagate their information to all other nodes through the edges after multiple layers. As a result, the representations of different nodes tend to be similar or indistinguishable to each other, influencing the performance of the GNN models. To overcome this drawback, the JK-Net~\cite{xu2018representation} has been developed to mitigate the over-smoothing problem by aggregating the outputs of all GNN layers. Another manner is to adequately capture the node features from the neighbor nodes at different distances, and this can be achieved by establishing the connections between nodes that are reachable within $n$ steps of a random walk, typical examples include the MixHop model~\cite{abu2019mixhop} and the N-GCN model~\cite{abu2020n}. However, the direction of the random walk does not always expand outward along the direction of the breadth-first search, thus the so-called distant neighbor information extracted from the two models may still contain a significant amount of redundant information from the nearby neighbors. Since the graph pooling operations are defined associated with the GNN models, the over-smoothing problem can in turn influence the effectiveness of existing graph pooling methods.

\begin{figure*}
  \centering
  \includegraphics[width=0.99\textwidth, trim=40 20 20 90, clip]{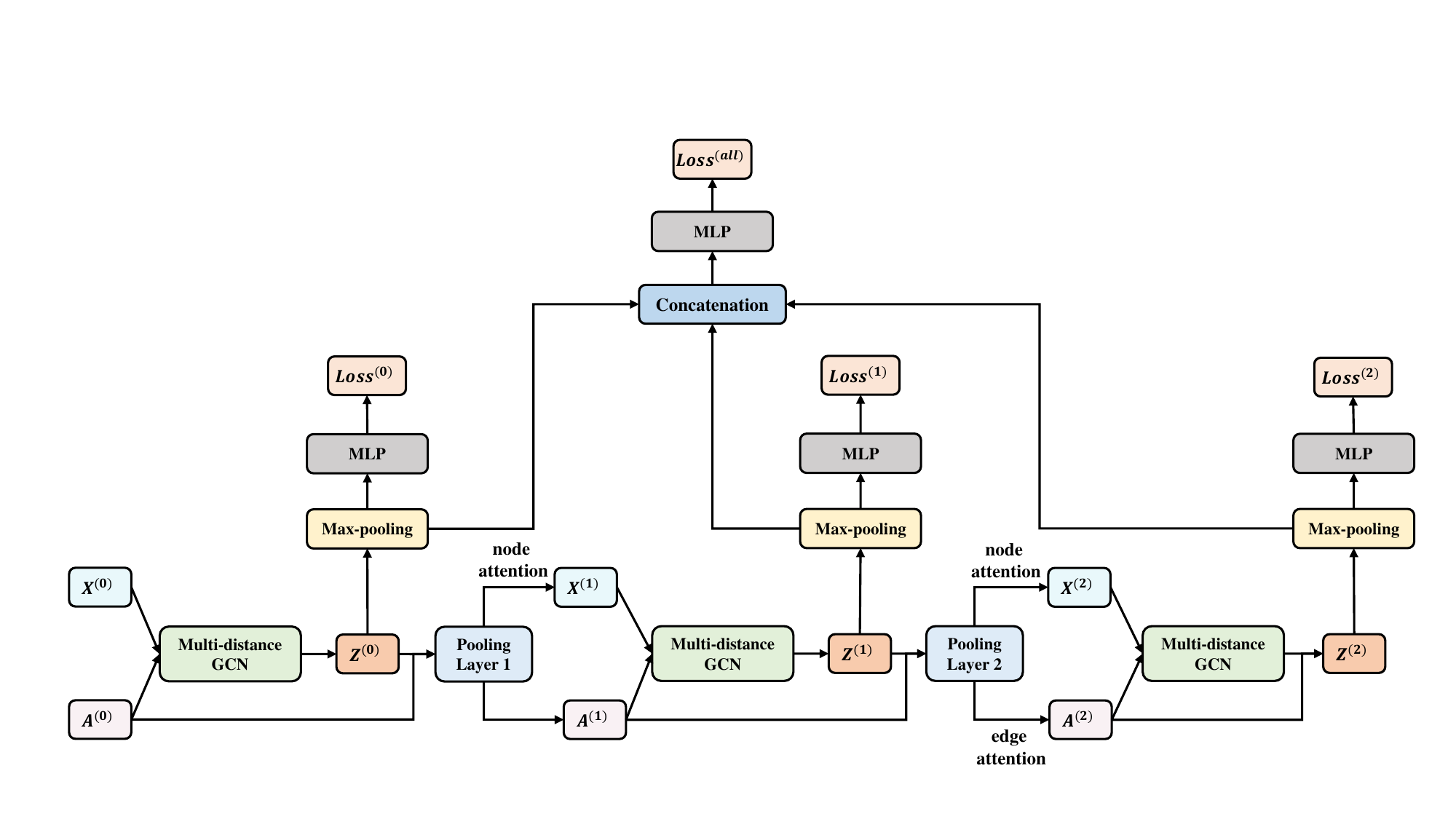}
  \vspace{-0pt}
  \caption{The architecture of the MD-GNN model associated with the proposed ENADPool.}
  \vspace{-0pt}
  \label{architecture}
  \vspace{-0pt}
\end{figure*}

The aim of this paper is to overcome the aforementioned shortcomings by defining a novel ENADPool operation associated with a new MD-GNN model. One innovation of the proposed ENADPoolis that it can consider the attention-based importance of different edges when the operation aggregates the edge connectivity strengths, enabling more selectively aggregated information from neighbors. The architecture of the proposed method is shown in Figure \ref{architecture}, and the main contributions are threefold.

\textbf{First}, we propose a cluster-based hierarchical ENADPool operation. Unlike the classical hierarchical pooling methods, the proposed ENADPool operation is defined based on the hard assignment, guaranteeing that each node can be assigned into an unique cluster and thus addressing the shortcoming of the node feature disruption. Moreover, unlike the classical attention-based hierarchical pooling methods that only assign different attentions to nodes, the proposed ENADPool operation can simultaneously identifies the importance of different nodes within each separated cluster and the edges between the clusters, addressing the shortcomings of the edge-node based structure information aggregation arising in the classical hierarchical pooling operation. To further address the computational expensive problem of existing attention-based hierarchical pooling methods, the proposed ENADPool operation directly adopts the self-attention mechanism to all nodes, and eliminates the need of the separated attention computations within each individual cluster. Moreover, we introduce an adjustment to the attention calculation process, and transform the dot product attention into the additive attention, reducing the number of the trainable parameters. \textbf{Second}, we propose a new Multi-distance GNN (MD-GNN) model associated with the proposed ENADPool operation. Since the MD-GNN model can propagate the feature information of the nodes at different distances with less redundancy. The proposed ENADPool operation associated with new MD-GNN model thus effectively mitigate the over-smoothing problem  arising in most existing GNN models. \textbf{Third}, we empirically evaluate the classification performance of the proposed ENADPool associated with the MD-GNN model. The experiments demonstrate the effectiveness of the proposed approach.

\section{The Literature Review of Related works}

In this section, we briefly review some existing works that are relate to this work, including the classical GNN models, graph pooling operations, and graph attention mechanisms.

\subsection{The Graph Convolutional Neural Networks}

The GNNs have powerful performance for graph data analysis. In early works, Bruna et al.~\cite{bruna2013spectral} have classified the GNNs into two categories, i.e., the spectral-based convolutional GNNs and the spatial-based convolutional GNNs. The spectral-based convolutional GNNs are based on the classical convolution operation in the spectral domain, that transforms the graphs into the spectral domain through the eigenvectors of the Laplacian matrix and perform the filter operation by multiplying the graph by a series of filter coefficients. To overcome the inefficiency of classical spectral-based GNNs, the ChebNet model~\cite{defferrard2016convolutional} has been developed and employs the rank-$k$ approximation of the Chebyshev polynomials as the filters and demonstrates the theoretical relationship between the spectral-based and spatial-based GNNs, i.e., the spectral-based GNNs can be seen as the one kind of the spatial-based GNNs. On the other hand, the spatial-based convolutional GNNs are similar to the classical convolution operation in the Euclidean space, that transforms the data within each local region. These spatial-based GNNs naturally incorporate the graph structure into node embeddings by aggregating the information between neighboring nodes. One popular representative methods is the GraphSAGE model~\cite{hamilton2017inductive} that samples a fixed number of neighboring nodes for each node and generate node embeddings through the inductive unsupervised learning.

Unfortunately, the above GNN models suffer from the notorious over-smoothing problem, that is caused by the node feature aggregation strategy between neighbor vertices. Since the GNN models can continuously propagate the repetitive information between adjacent nodes for multiple times, the resulting node embeddings tend to be similar to each other, influencing the performance of the GNN models. One way to overcome this problem is to directly propagate the information between the nodes with distant distances rather than the adjacent nodes. Typical GNN models based on this way include the MixHop model~\cite{abu2019mixhop}, the N-GCN model~\cite{abu2020n}, etc. Another challenging arising in these GNN models is how to extract graph-level characteristics based on the local structural information of the original graphs for graph-level classification. One important way to achieve this is to employ the graph pooling operations and downsize the graph structures~\cite{ying2018hierarchical}.

\subsection{The Graph Pooling Operation}

The graph pooling operations can be generally classified into two main categories, i.e., the global pooling methods and the hierarchical pooling methods. Specifically, the global pooling operations are defined by simply computing the sum, averaging, or the maximum value over the nodes of the whole graph structure. Since the global pooling operations tend to ignore the structural differences between the nodes, they cannot effectively learn the topological information residing on the local nodes. To overcome this problem, the cluster-based hierarchical pooling operations, that focus more on assigning local nodes into different clusters and gradually compressing the nodes into a family of coarsened nodes as the hierarchical structure, are developed for GNN models. The DiffPool~\cite{ying2018hierarchical} may be the first cluster-based hierarchical pooling method. Specifically, for each $l$-th layer the DiffPool first employs a specified GNN model to generate the node embedding matrix as $Z^{(l)} \in \mathbb{R}^{N_l \times d_{l+1}}$, and then generates the soft node assignment matrix based on $Z^{(l)}$ as $S^{(l)}\in \mathbb{R}^{N_l \times N_{l+1}}$, i.e.,
\begin{equation}
Z^{(l)}=\text{GNN}_{l, \text{embed}}(A^{(l)}, X^{(l)}),
\end{equation}
and
\begin{align}
S^{(l)}&=\text{softmax}(Z^{(l)})\\ \nonumber 
&=\text{softmax}(\text{GNN}_{l, \text{pool}}(A^{(l)}, X^{(l)})),\label{soft_assignment}
\end{align}
where $l$ represents the corresponding layer of the GNN model associated with the DiffPool, $X\in \mathbb{R}^{N_l \times d_{l}}$ is the original input node feature matrix, $A_l\in \mathbb{R}^{N_l \times N_l}$ is the adjacency matrix, $N_l$ is the number of nodes, and $d$ represents the dimension of the embedding. During the pooling process, the node features assigned to the same cluster and the edge connectivity strengths between two corresponding clusters are merged into coarsened graph structures by directly summing up them. The assignment matrix $S^{(l)}$ are used to form the new coarsened node feature matrix $X^{(l+1)} \in \mathbb{R}^{ N_{l+1}\times d}$ and the coarsened adjacency matrix $A^{(l+1)}\in \mathbb{R}^{N_{l+1}\times N_{l+1}}$ for the next $l+1$-th layer as
\begin{equation}
X^{(l+1)}={S^{(l)}}^T Z^{(l)},
\end{equation}
and
\begin{equation}
A^{(l+1)}={S^{(l)}}^T A^{(l)} S^{(l)}.
\end{equation}

To further consider the importance of different nodes belonging to the same cluster and extract more effective hierarchical coarsened graph structure information, the ABDPool operation~\cite{liu2022abdpool} has made further improvements based on the Diffpool and employs a dot product attention mechanism to calculate the self-attention score matrix $AT$ for each node cluster as
\begin{equation}
AT^{(l)}_{(c)}=\text{softmax}(\frac{QK^T}{\sqrt {d}}),
\end{equation}
where the matrices $Q \in \mathbb{R}^{N_c \times d}$ and $K \in \mathbb{R}^{N_c \times d}$ are obtained through two distinct linear transformations, and $c$ represents the $c$-th cluster. Then, the aggregated weights of the nodes within each cluster is determined by the sum of its correlation coefficients with other nodes, i.e.,
\begin{equation}
AS^{(l)}_{(c)}=\text{softmax}(I_{N_c}-\text{diag}(AT^{(l)}_{(c)})).
\end{equation}
Unfortunately, the ADBPool cannot identify the importance of the edge-based connectivity between node clusters, when the ADBPool generates the new coarsened adjacency matrix. 

\subsection{The Graph Attention Mechanism}
The attention mechanism~\cite{vaswani2017attention} has been widely used in GNN models, typical examples include the Graph Attention Networks (GAT)~\cite{velivckovic2017graph}, the Gated Attention Networks (GaAN)~\cite{zhang2018gaan}, etc. Specifically, the GaAN model utilizes the idea of the dot product attention to compute the node attention as
\begin{equation}
\alpha_{ij}=\langle\mathrm{FC}_q(\vec{h_i}),\mathrm{FC}_k(\vec{h_j})\rangle,
\end{equation}
where $\mathrm{FC}$ denotes a linear transformation function, different subscripts of $\mathrm{FC}$ mean different transformation parameters, $\vec{h_i}$ represents the vectorial feature of the $i$-node. On the other hand, the GAT model employs the idea of the additive attention to calculate node attentions
\begin{equation}
\alpha_{ij}=\text{LeakyReLU}\left(\vec{\mathbf{a}}^T[\textbf{W}\vec{h_i}\|\textbf{W}\vec{h_j}]\right),
\end{equation}
where $\mathbf{W} \in \mathbb{R}^{{F}' \times{F}}$ is a weight matrix, and $\vec{\mathbf{a}} \in \mathbb{R}^{{2F}'}$ is a weight vector. For each $i$-th node, both the GaAN and GAT models can assign its $j$-th neighbor nodes (include the $i$-th node itself, i.e., $i\in j$) with different weights based on the node attentions.

\section{The Proposed Methods}
In this section, we propose a novel cluster-based hierarchical ENADPool operation. Moreover, we develop a new MD-GNN model associated with the ENADPool operation.

\subsection{The Proposed ENADPool Operation}

We define the proposed cluster-based hierarchical ENADPool operation, that can simultaneously identify the importance of different nodes in each cluster and different edges between corresponding clusters based on the attention mechanism after each pooling step. The computational architecture of \emph{\textbf{the proposed ENADPool for each $l$-layer}} mainly consists of three steps. \textbf{First}, for the input graph $G^{(l)}(V^{(l)},E^{(l)})$ of each $l$-layer, we assign the set of nodes $V^{(l)}$ ($N_l=|V^{(l)}|$) into different separated $N_{l+1}$ clusters, resulting in a hard node assignment matrix $S^{(l)}_\mathrm{H} \in \mathbb{R}^{N_l \times N_{l+1}}$. \textbf{Second}, we assign different $i$-th nodes belonging to the same cluster with different weights $\alpha_i^{(l)}$ through the node attention. Based on the node attention, we compute the weighted compressed node feature matrix $X^{(l+1)}\in \mathbb{R}^{N_{l+1} \times d_{l+1}}$ of the resulting coarsened graph $G^{(l+1)}$ for the next $l+1$-th layer. \textbf{Third}, we assign different $(i,j)$-th edges between the $i$-th and $j$-th nodes in the $p$-th and $q$-th node clusters respectively, associated with different weights $\beta_{ij}^{(l)}$ through the edge attention. Based on the edge attention, we compute the compressed adjacency matrix $A^{(l+1)} \in \mathbb{R}^{N_{l+1} \times N_{l+1}}$ (i.e., the weighted compressed edge connectivity) of the coarsened graph $G^{(l+1)}$ for the next $l+1$-th layer. Details of the three steps are defined as follows.

\textbf{Definition 1 (The Hard Node Assignment).} The hard node assignment can deterministically assign each node in $V^{(l)}$ of $G^{(l)}$ into an unique cluster, and the hard assignment matrix $S^{(l)}_\mathrm{H}\in \mathbb{R}^{N_l \times N_{l+1}}$ can be computed based on the soft assignment matrix $S^{(l)}\in \mathbb{R}^{N_l \times N_{l+1}}$ defined by Eq.(\ref{soft_assignment}), i.e.,
\begin{align}\label{eq:Sl}
S^{(l)}_\mathrm{H}&=\text{onehot}[S^{(l)}]\\ \nonumber
&=\text{onehot}[\text{softmax}(Z^{(l)})]\\ \nonumber
&=\text{onehot}[\text{softmax}(\text{GNN}_{l,\text{pool}}(A^{(l)}, X^{(l)}))],
\end{align}
where the operation $\emph{\textbf{onehot}}$ selects the maximum value of each row in $S^{(l)}$ as $1$ and others as $0$.

\begin{figure*}
  \centering
  \includegraphics[width=0.99\textwidth, trim=50 30 30 45, clip]{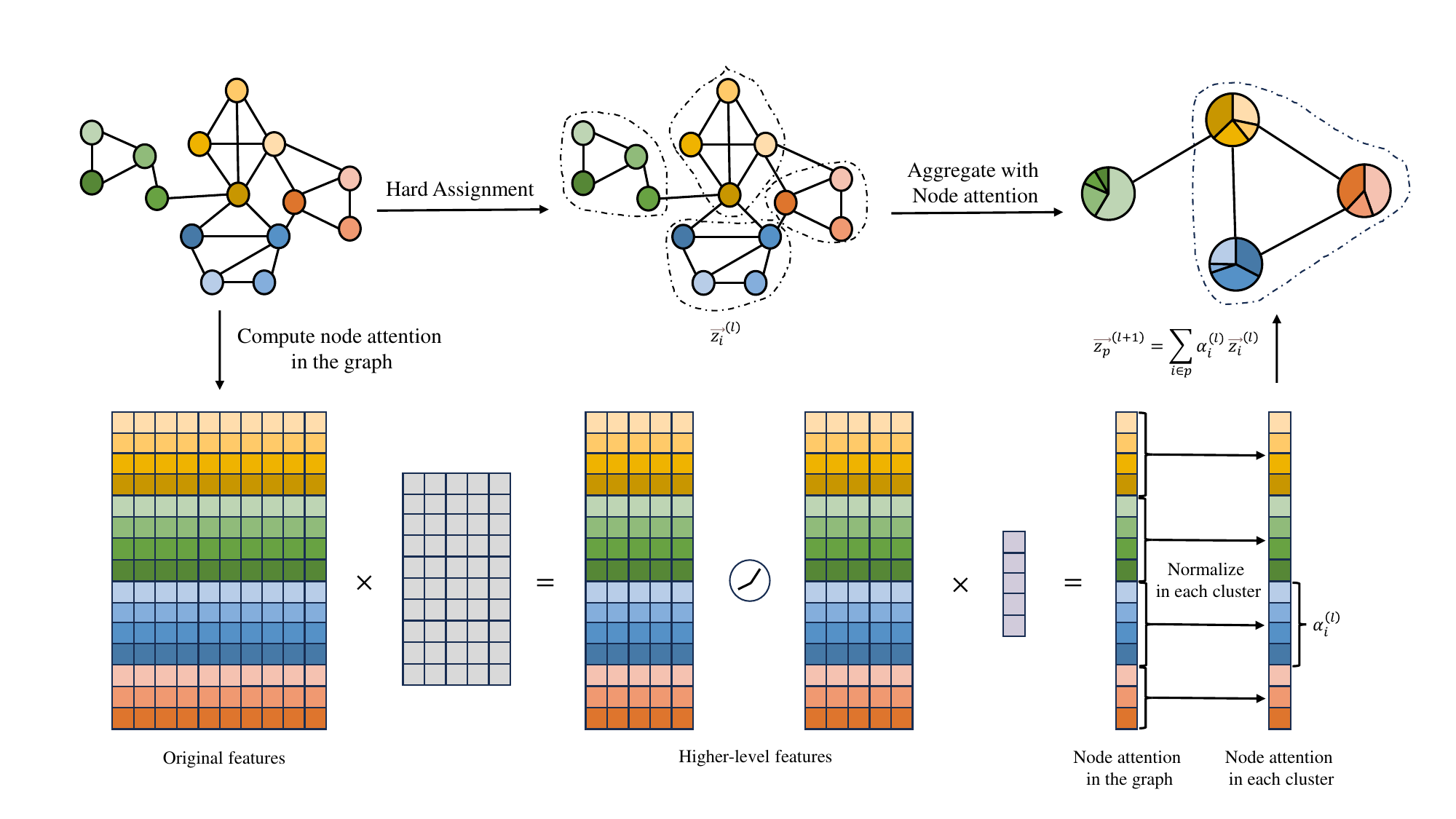}
  \caption{The process of aggregating the nodes for the coarsened graph through node attentions.}\label{node}
  \vspace{-0pt}
\end{figure*}

\textbf{Definition 2 (The Node-based Attention).} The process of this computational step is shown in Figure~\ref{node}. Specifically, we propose to compute the attention-based importance of different nodes over the entire graph $G^{(l)}$, and then normalize the importance of the nodes assigned to each individual cluster. The input is a set of node embeddings $Z^{(l)}=\{\vec{z_1}^{(l)},\vec{z_2}^{(l)},...,\vec{z_{N_l}}^{(l)}\}$ computed by a specified GNN model in Eq.(\ref{eq:Sl}), where $ \vec{z_i}^{(l)} \in \mathbb{R}^{d_{l+1}}$ is the embedding vector of the $i$-th node, $N_l$ is the number of nodes and $d_l$ is the dimension of each node embedding for the $l$-th layer. To enhance the understanding of the node relationships, we commence by transforming the input features $Z^{(l)}$ into more meaningful representations. This is achieved by applying a linear transformation to each node, parameterized by the weight matrix $\mathbf{W}_m^{(l)} \in \mathbb{R}^{{d_{l+1}}' \times{d_{l+1}}}$, where ${d_{l+1}}'$ denotes the dimensionality of the transformed feature. To introduce the nonlinearity, the transformed high-level features are then passed through a $\text{LeakyReLU}$ activation function. Subsequently, the self-attention mechanism is employed to calculate the attention coefficients for all nodes. In this paper, this mechanism can be implemented using a fully connected layer, which can be parameterized as $\vec{\mathbf{a}}_m^{(l)} \in \mathbb{R}^{{d_{l+1}}'}$. Thus, the importance of the $i$-th node in the whole graph can be calculated as 
\begin{equation}
\label{eq:mi}
m_i^{(l)} = {\vec{\mathbf{a}}^{(l)}_{m}}{}^{T} \mathrm{LeakyReLU}(\textbf{W}^{(l)}_m\vec{z_i}^{(l)}).
\end{equation}
In order to aggregate the nodes belonging to each cluster, we need to normalize the attention-based importance of all nodes assigned to the same cluster. This can be accomplished by applying the $\mathrm{softmax}$ function, and the specific calculation process is defined as
\begin{align}\label{eq:ai}
\alpha_i^{(l)} & = \mathrm{softmax}_p(m_i^{(l)}) \\ \nonumber
&=\frac{\exp\left({\vec{\mathbf{a}}^{(l)}_m}{}^T \mathrm{LeakyReLU}(\textbf{W}^{(l)}_m\vec{z_i}^{(l)})\right)}{\sum_{j \in p}\exp\left({\vec{\mathbf{a}}^{(l)}_m}{}^T \mathrm{LeakyReLU}(\textbf{W}^{(l)}_m\vec{z_j}^{(l)})\right)},
\end{align}
where $p$ denotes the $p$-th cluster, $i$ represents the $i$-th node assigned to the cluster, and $j\in p$ represents all the $j$-th nodes (i.e., $i\in j$) assigned to the cluster. Finally, we compress the nodes of each $p$-th cluster as the $p$-th coarsened node of the coarsened graph $G^{(l+1)}$ for the next $l+1$-th layer, by computing the weighted sum $\sum_{i \in p}\alpha_i^{(l)}\vec{z_i}^{(l)}$ of these nodes associated with their attention-based importance coefficients $\alpha_i^{(l)}$. The resulting node feature matrix $X^{(l+1)}\in \mathbb{R}^{N_{l+1} \times d_l}$ of $G^{(l+1)}$ can be computed by row-wisely concatenating all the coarsened node embeddings as
\begin{equation}
\label{eq:zp}
X^{(l+1)}= \mathrm{CONCATE}_{p=1}^{n_{l+1}} [\sum_{i \in p}\alpha_i^{(l)}\vec{z_i}^{(l)}].
\end{equation}

\begin{figure*}
  \centering
  \vspace{-0pt}
  \includegraphics[width=1\textwidth, trim=100 120 100 120, clip]{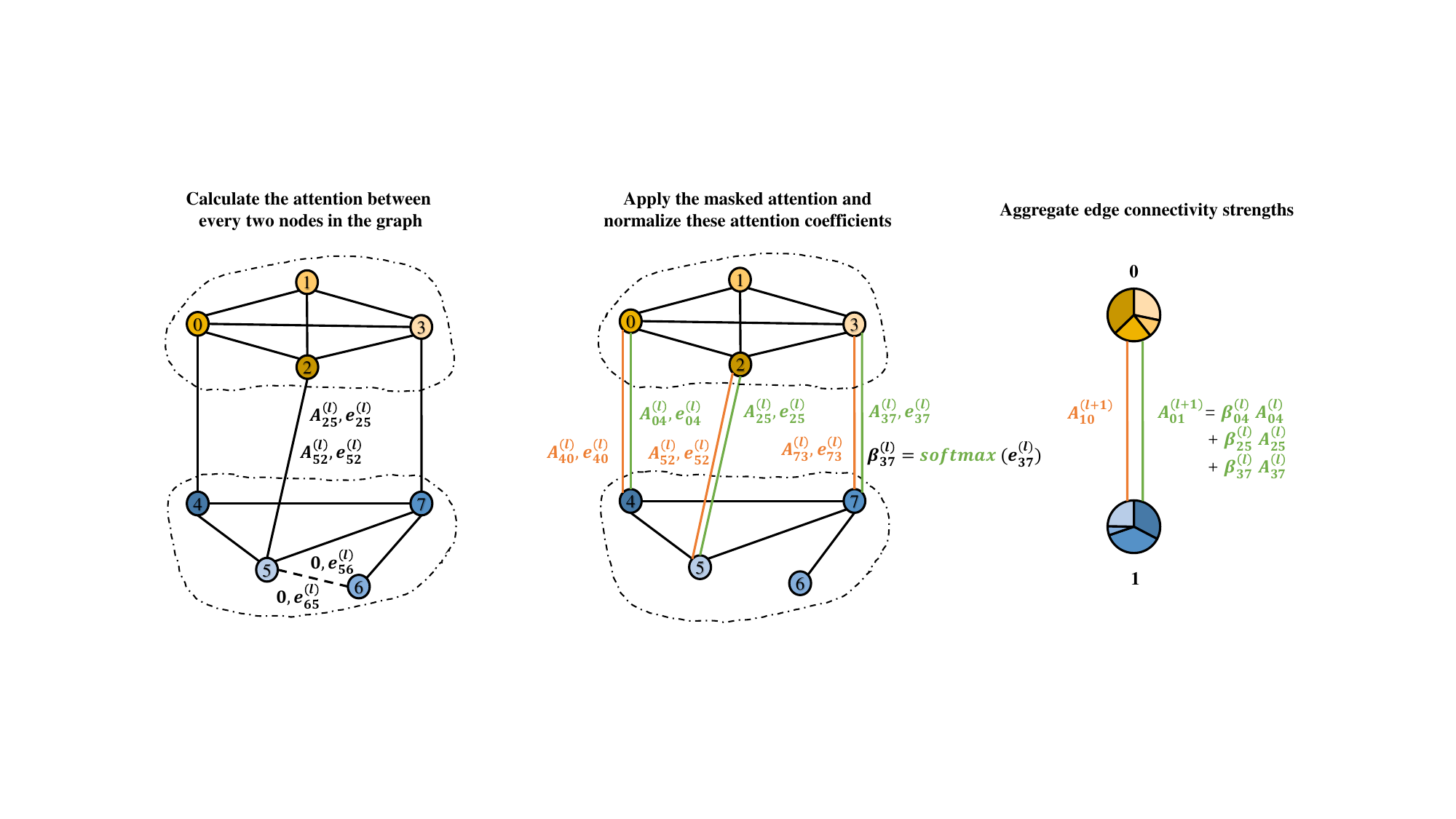}
  \caption{The process of aggregating the edges for the coarsened graph through edge attentions.}\label{edge}
  \vspace{-0pt}
\end{figure*}

\textbf{Definition 3 (The Edge-based Attention).} The process of this computational step is shown in Figure~\ref{edge}. Specifically, we propose to compute the attention-based importance for the edges between the node clusters. We first utilize a weight matrix $\mathbf{W}_e^{(l)} \in \mathbb{R}^{{d_l}' \times {d_l}}$ and a self-attention mechanism $a_e^{(l)}$ to compute the initial attention of the edge between the $i$-th and $j$-th nodes as
\begin{equation}
\label{eq:sij}
s_{ij}^{(l)} = a_e^{(l)}(\mathbf{W}_e^{(l)}\vec{z_i}^{(l)}, \mathbf{W}_e^{(l)}\vec{z_j}^{(l)})
\end{equation}
Moreover, we employ the masked attention mechanism to merely consider the edges between the same pair of node clusters, and normalize their importance using the $\mathrm{softmax}$ function as
\begin{equation}
\label{eq:bij}
\beta_{ij}^{(l)} = \mathrm{softmax}_{pq}(e_{ij}^{(l)})=\frac{\mathrm{exp}(e_{ij}^{(l)})}{\sum_{x \in p,y \in q} {\mathrm{exp}(e_{xy}^{(l)})}},
\end{equation}
where the $i$-th and $j$-th nodes belong to the $p$-th and $q$-th node clusters respectively. Since the attention mechanism $a_e^{(l)}$ can be implemented as a fully connected layer, parameterized by a weight vector $\vec{\mathbf{a}}_e^{(l)} \in \mathbb{R}^{{2d_l}'}$. The resulting edge attention coefficients between each pair of clusters can be defined as
\begin{equation}\label{eq:bij2}
\small{\beta_{ij}^{(l)}=\frac{\exp\left(\mathrm{LeakyReLU}\left({\vec{\mathbf{a}}^{(l)}_e}{}^T[\textbf{W}^{(l)}_e\vec{z_i}^{(l)}\|\textbf{W}^{(l)}_e\vec{z_j}^{(l)}]\right)\right)}{\sum_{x \in p,y \in q}\exp\left(\mathrm{LeakyReLU}\left({\vec{\mathbf{a}}^{(l)}_e}{}^T[\textbf{W}^{(l)}_e\vec{z_x}^{(l)}\|\textbf{W}^{(l)}_e\vec{z_y}^{(l)}]\right)\right)}}
\end{equation}
where the || represents the concatenation operation. Finally, we utilize these edge attention coefficients as the weights to aggregate the edge connectivity strengths between different clusters as the compressed adjacency matrix $A^{(l+1)}$ of the coarsened graph $G^{(l)}$ for the next $l+1$-th layer, i.e.,
\begin{equation}
\label{eq:Al}
A^{(l+1)}={S^{(l)}}^T (E^{(l)} \odot A^{(l)}) S^{(l)}
\end{equation}
where the matrix $E^{(l)} \in \mathbb{R}^{{N_l} \times{N_l}}$ encapsulates the edge attention coefficients, and $E^{(l)}_{ij}= \beta_{ij}^{(l)}$.

\subsection{The MD-GNN Model associated with the Proposed ENADPool}

\begin{figure*}
  \centering
   \vspace{-0pt}
  \includegraphics[width=0.99\textwidth, trim=100 150 100 150, clip]{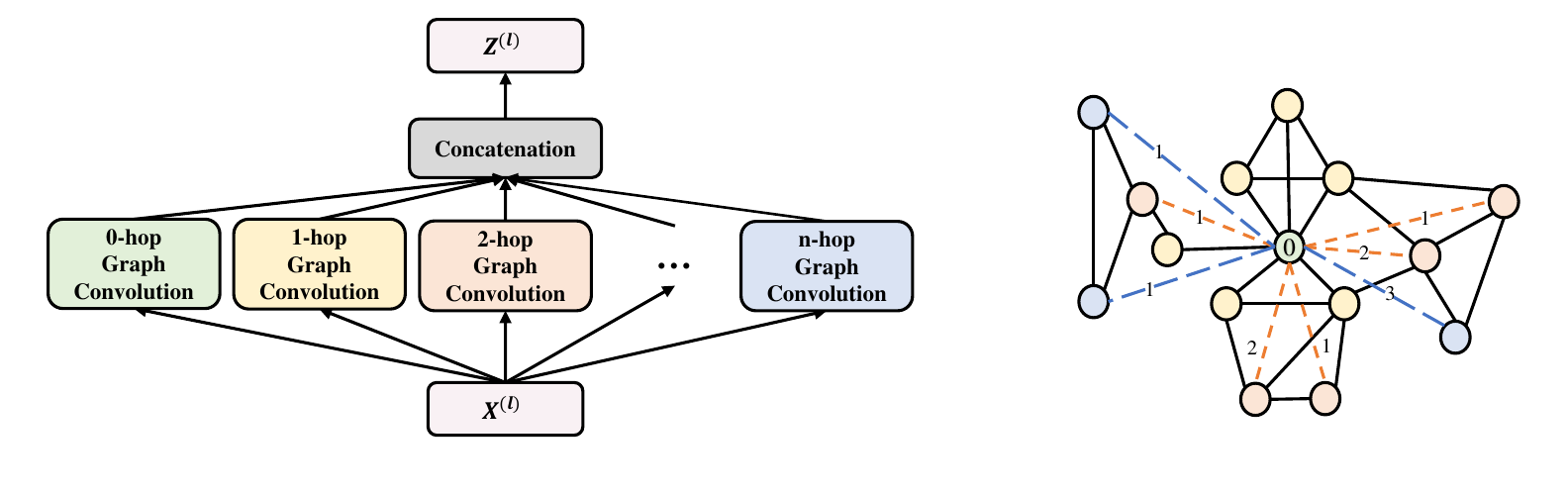}
  \vspace{-0pt}
  \caption{\textbf{Left}: The architecture of the MD-GNN. \textbf{Right}: An illustration of the MD convolution process, the different colors denote the nodes that participate the convolution operation with node $0$ and can arrive in node $0$ with different steps, e.g., the 1, 2 and 3 steps for this instance.}\label{GCN}
  \vspace{-0pt}
\end{figure*}

The graph pooling operations are defined associated with the GNN models, since the operations need the node embeddings extracted from the GNNs as the input node features. To reduce the notorious over-smoothing problem arising in existing GNNs and further improve the effectiveness of the proposed ENADPool, we develop a new MD-GNN model that directly aggregates the node information of different distances (i.e., different random walk steps), by reconstructing the graph topology structure and providing a comprehensive view of the structure information around each node. This is inspired by the strategy~\cite{abu2019mixhop} of adequately capturing the node features from different $h$-hop neighbor nodes, and the architecture of the MD-GNN is illustrated in Figure~\ref{GCN}. Specifically, for the input graph $G^{(l)}(V^{(l)},E^{(l)})$ with its adjacency matrix $A^{(l)} \in \mathbb{R}^{N_l \times N_l}$ ($|V^{(L)}|=N_l$), we employ a mask mechanism to filter the connections in ${A^{(l)}}^h \in \mathbb{R}^{N_l \times N_l}$. To this end, we commence by computing a family of binary matrices $R_h\in \{0,1\}^{N_l \times N_l}$, and the $(i,j)$-th entry $R_h(i,j)$ satisfies
\begin{equation}
R_h(i,j)=\left\{
\begin{array}{cl}
1   & \mathrm{if} \  {A^{(l)}}^h (i,j)>0; \\
0   & \mathrm{otherwise}.
\end{array} \right.
\label{RMatrix}
\end{equation}
Here, $h$ varies form $0$ to $H$, and \textbf{$R_h(i,j)=1$ indicate that the random walk departing from the $i$-th node can arrive in the $j$-th node within $h$ walk steps, i.e., may also include less than $h$ steps}.  Moreover, we compute another family of binary matrices $U_h\in \{0,1\}^{N_l \times N_l}$ as
\begin{equation}
\label{eq:Lx}
U_h = \text{clip}[R_h-\Sigma_{x=0}^{h-1}{R_x}],
\end{equation}
where the function \textbf{clip(*)} is used to restrict the range of the values in $U_h$ to be between $0$ and $1$. Here, \textbf{$U_h(i,j)=1$ indicates that the walk step value of the random walk departing from the $i$-th node and arriving in the $j$-th node is exactly $h$ or at least $h$}. In other words, $U_h(i,j)$ indicates the $h$-hop neighborhood nodes indexed by $j$ rooted at each node indexed by $i$, in terms of $h$ random walk steps. With $U_h$ to hand, we construct the new topology structures $T_h$ for $G^{(l)}$ by employing $U_h$ as the mask for ${A^{(l)}}^h$, i.e.,
\begin{equation}
\label{eq:Tx}
T_h = U_h \odot {A^{(l)}}^h.
\end{equation}
With $T_h$ ($0\leq h \leq H$) to hand, the MD-GNN model can be defined in the terms of the node feature matrix $X^{(l)}$ and the adjacency matrix $A^{(l)}$ as
\begin{align}\label{eq:Zl}
\text{MD-GNN}({A^{(l)}}^h, X^{(l)})&=\text{GCN}(T_0, X^{(l)})||\text{GCN}(T_1, X^{(l)})|| \\ \nonumber
& \ldots||\text{GCN}(T_H, X^{(l)}).
\end{align}

\section{Experiments}

We empirically compare the proposed method with other graph deep learning approaches for graph classification on benchmark datasets, including the D\&D~\cite{dobson2003distinguishing}, PROTEINS~\cite{borgwardt2005protein}, NCI1/NCI109~\cite{wale2008comparison}, FRANKENSTEIN \cite{orsini2015graph}, and REDDIT-B~\cite{yanardag2015deep} datasets. Detailed statistics of these datasets are shown in Table~\ref{dataset}. Note that, if the graphs do not have node labels, the node degree can be used as the label. 

\begin{table*}
  \caption{Dataset statistics.}
  \label{dataset}
  \centering
  \footnotesize
  \begin{tabular}{lcccc}
    \hline
    \textbf{Dataset}     & \textbf{Graphs}     & \textbf{Classes}      & \textbf{Nodes(mean)}      & \textbf{Edges(mean)}  \\
    \hline
    D\&D     & 1178     & 2      & 284.32      & 715.66  \\
    NCI1     & 4110     & 2      & 29.87      & 32.30  \\
    NCI109     & 4127     & 2      & 29.68      & 32.13  \\
    PROTEINS     & 1113     & 2      & 39.06      & 72.82  \\
    FRANKENSTEIN     & 4337     & 2      & 16.90      & 17.88  \\
    REDDIT-B     & 2000     & 2      & 429.63      & 497.75  \\
    \hline
  \end{tabular}
  \vspace{-0pt}
\end{table*}

\begin{table*}
\vspace{-0pt}
\footnotesize
  \caption{Classification accuracy (in \% $\pm$ standard error) for comparisons. ENADPool(N) indicates that the node attention is adopted, ENADPool(E) indicates that the edge attention is adopted, and ENADPool(NE) indicates that both the node and edge attentions are adopted. OOR means Out of Resources, either time (>72h) or GPU memory. - means that it was not evaluated in original papers.}
  \label{result}
  \centering
  \begin{tabular}{p{0.2cm}lcccccc}
    \hline
    &      & \textbf{D\&D}     & \textbf{NCI1}      & \textbf{NCI109}      & \textbf{PROTEINS}      & \textbf{FRANKENSTEIN}      & \textbf{REDDIT-B}  \\
    \hline
    \multirow{6}{*}{\rotatebox[origin=c]{90}{\centering Baselines}}
    & DGCNN     & 79.37 $\pm$ 0.94     & 74.44 $\pm$ 0.47      & 69.79 $\pm$ 0.25      & 75.54 $\pm$ 0.94      & 63.44 $\pm$ 0.65      & 76.02 $\pm$ 1.73  \\

    & GCKN     & 77.3 $\pm$ 4.0     & 79.2 $\pm$ 1.2      & -      & 76.1 $\pm$ 2.8      & -      & OOR  \\
    & KerGNN   & 78.9 $\pm$ 3.5     & 76.3 $\pm$ 2.6      & -      & 75.5 $\pm$ 4.6      & -      & 82.0$\pm$ 2.5  \\

    & SAGPool(G)      & 76.19 $\pm$ 0.94     & 74.18 $\pm$ 1.20     & 74.06 $\pm$ 0.78     & 70.04 $\pm$ 1.47      & 62.57 $\pm$ 0.60      & 74.45 $\pm$ 1.73  \\
    & SAGPool(H)      & 76.45 $\pm$ 0.97     & 67.45 $\pm$ 1.11     & 67.86 $\pm$ 1.41     & 71.86 $\pm$ 0.97      & 61.73 $\pm$ 0.76      & 75.53 $\pm$ 3.53  \\
    & ASAP      & 76.87 $\pm$ 0.70     & 71.48 $\pm$ 0.42      & 70.07 $\pm$ 0.55      & 74.19 $\pm$ 0.79      & 66.26 $\pm$ 0.47      & OOR  \\
    & DiffPool      & 77.33 $\pm$ 0.43     & 77.08 $\pm$ 0.38      & 75.70 $\pm$ 0.97      & 76.72 $\pm$ 0.25      & 63.80 $\pm$ 0.04      & 82.12 $\pm$ 1.06  \\
    & ABDPool      & 74.13 $\pm$ 0.52     & OOR      & OOR      & 73.24 $\pm$ 0.91      & OOR      & OOR  \\
    \hline
    \multirow{3}{*}{\rotatebox[origin=c]{90}{\centering Ours}}
    & ENADPool(N)      & \textbf{79.50 $\pm$ 0.52}     & \textbf{79.61 $\pm$ 0.24}      & \textbf{78.61 $\pm$ 0.29}      & \textbf{77.44 $\pm$ 0.10}      & \textbf{67.59 $\pm$ 0.33}      & \textbf{88.21 $\pm$ 0.51}  \\
    & ENADPool(E)      & \textbf{80.04 $\pm$ 0.29}     & \textbf{79.44 $\pm$ 0.20}      & \textbf{78.31 $\pm$ 0.34}      & \textbf{77.21 $\pm$ 0.5}1      & \textbf{67.51 $\pm$ 0.09 }     & \textbf{88.38 $\pm$ 0.41}  \\
    & ENADPool(NE)      & \textbf{79.45 $\pm$ 0.14}     & \textbf{79.34 $\pm$ 0.31}      & \textbf{78.67 $\pm$ 0.19}      & \textbf{77.12 $\pm$ 0.15}      & \textbf{67.68 $\pm$ 0.13}      & \textbf{88.46 $\pm$ 0.28}  \\
    \hline
  \end{tabular}
  \vspace{-0pt}
\end{table*}

\subsection{Baselines and Experimental Settings}

We adopt four hierarchical pooling methods as baselines for comparisons, including the Self-Attention Graph Pooling with the hierarchical pooling (SAGPool(H))~\cite{lee2019self}, the Adaptive Structure Aware Pooling (ASAPool)~\cite{ranjan2020asap}, the Differentiable Pooling (DiffPool)~\cite{ying2018hierarchical}, and the Attention-based Differentiable Pooling (ABDPool)~\cite{liu2022abdpool}. Moreover, we also consider several global pooling methods for comparisons, including the Deep Graph Convolutional Neural Network (DGCNN)~\cite{zhang2018end}, the SAGPool with the global structure (SAGPool(G))~\cite{lee2019self}, the Graph Kernel-based GNN (KerGNN)~\cite{DBLP:conf/aaai/FengY0T22}, and the Graph Convolutional Kernel Network (GCKN)~\cite{DBLP:conf/icml/ChenJM20}. For the proposed method, we employ 10-fold cross-validation for the model assessment with a 90\%/10\% training/validation split, and report the average accuracy and standard deviation of 10 runs. For these baselines, we directly report the results from their original papers, since the evaluations of these methods followed the same setting of ours.


  \begin{table*}
    \centering
    \vspace{-0pt}
    \caption{Classification Accuracy (In \% $\pm$ Standard Error) for validating the impact of each module: hard cluster-based assignment.}\label{ablationa}
    \vspace{-0pt}
    \begin{tabular}{ccc}
      \hline
      \textbf{Assignment method} & \textbf{PROTEINS} & \textbf{D\&D} \\
      \hline
      Soft clustering & 76.72 $\pm$ 0.25 & 77.33 $\pm$ 0.43 \\
      Hard clustering & \textbf{76.80 $\pm$ 0.09} & \textbf{78.34 $\pm$ 0.62} \\
     \hline
    \end{tabular}
  \end{table*}

  \begin{table*}
    \centering
    \vspace{-0pt}
    \caption{MD-GNN (Classification Accuracy (In \% $\pm$ Standard Error) for validating the impact of each module: on the basis of hard cluster-based assignment).}\label{ablationb}
    \vspace{-0pt}
    \begin{tabular}{ccc}
      \hline
      \textbf{GNN architecture} & \textbf{PROTEINS} & \textbf{D\&D} \\
      \hline
      JK-Net & 76.80 $\pm$ 0.09 & 78.34 $\pm$ 0.62 \\
      MD-GNN & \textbf{76.92 $\pm$ 0.41} & \textbf{79.20 $\pm$ 0.48} \\
      \hline
    \end{tabular}
  \end{table*}

  \begin{table*}
    \centering
    \vspace{-0pt}
    \caption{Attention mechanisms on nodes and edges (Classification Accuracy (In \% $\pm$ Standard Error) for validating the impact of each module: on the basis of hard cluster-based assignment and MD-GNN).}\label{ablationc}
    \vspace{-0pt}
    \begin{tabular}{cccc}
      \hline
      \textbf{Node-based attention} & \textbf{Edge-based attention} & \textbf{PROTEINS} & \textbf{D\&D} \\
      \hline
      $\times$ & $\times$ & 76.92 $\pm$ 0.41 & 79.20 $\pm$ 0.48 \\
      \checkmark & $\times$ & \textbf{77.44 $\pm$ 0.10} & 79.50 $\pm$ 0.52 \\
      $\times$ & \checkmark & 77.21 $\pm$ 0.51 & \textbf{80.04 $\pm$ 0.29} \\
      \checkmark & \checkmark & 77.12 $\pm$ 0.15 & 79.45 $\pm$ 0.14 \\
      \hline
    \end{tabular}
  \end{table*}

For the proposed method, we employ the same architecture for all datasets, and the only hyperparameters that need to be optimized are the number of epochs and the batchsize for the mini-batch gradient descent algorithm. We set the initial size of the graph as $512$, and employ 2 pooling layers, where the number of the clusters for the next layer is set as 25\% of that for the current layer. To compute the embedding matrix and the node assignment matrix, we utilize two separated MD-GNN models, where the number of graph convolution operations is uniformly set as $6$, and each operation has an output dimension of $32$. During the graph convolution process, we employ the ReLU~\cite{nair2010rectified} as the nonlinear activation function, and maintain a dropout rate of $0.25$. We utilize a fully connected neural network with one hidden layer for classification, the hidden layer consists of $256$ hidden units and incorporates a dropout ratio of $0.5$. Furthermore, inspired by the GoogLeNet~\cite{szegedy2015going}, for the training process we also employ the auxiliary classifiers. This involves additional fully connected neural networks to predict labels in the middle layers. These networks have one hidden layer with $128$ hidden units, and the dropout ratio is $0.5$. During the training process, the losses computed by the auxiliary classifiers are incorporated into the total loss with a smaller weight ($\omega = 0.2$) to assist in training. However, during the testing phase, the auxiliary classifiers are discarded. These classifiers contribute to propagate gradient and prevent gradient vanishing. Finally, for the training process of the proposed method, we utilize the Adam optimizer~\cite{kingma2014adam} with a weight decay of $0.001$ and a learning rate of $0.00001$. To prevent the over-fitting, we employ the early stopping strategy with a patience parameter $10$, i.e., we halt the training if the validation accuracy does not improve for $10$ epochs.

\subsection{Results and Discussions}

Table~\ref{result} indicates that the proposed method can significantly outperform the alternative methods, and the reasons for the effectiveness are fourfold. \textbf{First}, unlike existing cluster-based hierarchical pooling operations, the proposed ENADPool is defined based on the hard node assignment, addressing the shortcoming of the node feature disruption. \textbf{Second}, only the proposed ENADPool operation can simultaneously identifies the attention-based importance of different nodes assigned into each separated cluster and the corresponding edges between the clusters, addressing the shortcomings of the uniform edge-node based structure information aggregation arising in the classical hierarchical pooling operation. \textbf{Third}, the proposed ENADPool operation is associated with the new MD-GNN model, that can reduce the over-smoothing problem by propagating the node feature information at different distances, learning more effective graph representations. \textbf{Fourth}, the ENADPool operation can address the shortcoming of the existing GNNs associated with the global pooling, that neglects the feature distribution associated with different nodes.

\subsection{Ablation Experiments}
To further evaluate the proposed ENADPool with different modules, we perform ablation experiments on the PROTEINS and D\&D datasets. Specifically, Table~\ref{ablationa} verifies the effectiveness of the proposed ENADPool with the hard assignment, and it can be observed that changing the assignment method resulted in significant improvements in accuracies. Table~\ref{ablationb} is utilized to assess the effectiveness of the proposed ENADPool with the new MD-GNN. We modify the GNN architecture based on the hard assignment, transitioning from the JK-Net to the MD-GNN. Both architectures are configured to capture the graph structure information within 6 hops. Clearly, the architecture employing the MD-GNN achieves higher accuracies, indicating that the MD-GNN can improve the effectiveness of the proposed ENADPool. Table~\ref{ablationc} is utilized to assess the effectiveness of the attention mechanisms on nodes and edges, by employing the hard assignment and the MD-GNN. Clearly, each attention mechanism is effective. However, due to the distinct data distributions of each dataset, there are slight variations when different attention mechanisms are employed.

\section{Conclusions}

In this paper, we have proposed a novel cluster-based hierarchical ENADPool operation associated with a new MD-GNN model for graph classification. The proposed method can compress the node features as well as their edge connectivity strengths into the resulting hierarchical structure based on the attention mechanism after each pooling step, significantly identifying the importance of the nodes and edges to form the coarsened graph structures. Moreover, the proposed method can reduce the over-smoothing problem arising in existing GNNs. Experiments demonstrate the effectiveness. 




\balance


\bibliographystyle{IEEEtran}
\bibliography{ref}

\begin{thebibliography}{10}
\providecommand{\url}[1]{#1}
\csname url@samestyle\endcsname
\providecommand{\newblock}{\relax}
\providecommand{\bibinfo}[2]{#2}
\providecommand{\BIBentrySTDinterwordspacing}{\spaceskip=0pt\relax}
\providecommand{\BIBentryALTinterwordstretchfactor}{4}
\providecommand{\BIBentryALTinterwordspacing}{\spaceskip=\fontdimen2\font plus
\BIBentryALTinterwordstretchfactor\fontdimen3\font minus
  \fontdimen4\font\relax}
\providecommand{\BIBforeignlanguage}[2]{{%
\expandafter\ifx\csname l@#1\endcsname\relax
\typeout{** WARNING: IEEEtran.bst: No hyphenation pattern has been}%
\typeout{** loaded for the language `#1'. Using the pattern for}%
\typeout{** the default language instead.}%
\else
\language=\csname l@#1\endcsname
\fi
#2}}
\providecommand{\BIBdecl}{\relax}
\BIBdecl

\bibitem{xu2018powerful}
K.~Xu, W.~Hu, J.~Leskovec, and S.~Jegelka, ``How powerful are graph neural
  networks?'' \emph{arXiv preprint arXiv:1810.00826}, 2018.

\bibitem{wu2022graph}
S.~Wu, F.~Sun, W.~Zhang, X.~Xie, and B.~Cui, ``Graph neural networks in
  recommender systems: a survey,'' \emph{ACM Computing Surveys}, vol.~55,
  no.~5, pp. 1--37, 2022.

\bibitem{li2017diffusion}
Y.~Li, R.~Yu, C.~Shahabi, and Y.~Liu, ``Diffusion convolutional recurrent
  neural network: Data-driven traffic forecasting,'' \emph{arXiv preprint
  arXiv:1707.01926}, 2017.

\bibitem{sun2020graph}
M.~Sun, S.~Zhao, C.~Gilvary, O.~Elemento, J.~Zhou, and F.~Wang, ``Graph
  convolutional networks for computational drug development and discovery,''
  \emph{Briefings in bioinformatics}, vol.~21, no.~3, pp. 919--935, 2020.

\bibitem{kipf2016semi}
T.~N. Kipf and M.~Welling, ``Semi-supervised classification with graph
  convolutional networks,'' \emph{arXiv preprint arXiv:1609.02907}, 2016.

\bibitem{hamilton2017inductive}
W.~Hamilton, Z.~Ying, and J.~Leskovec, ``Inductive representation learning on
  large graphs,'' \emph{Advances in neural information processing systems},
  vol.~30, 2017.

\bibitem{velivckovic2017graph}
P.~Veli{\v{c}}kovi{\'c}, G.~Cucurull, A.~Casanova, A.~Romero, P.~Lio, and
  Y.~Bengio, ``Graph attention networks,'' \emph{arXiv preprint
  arXiv:1710.10903}, 2017.

\bibitem{grover2016node2vec}
A.~Grover and J.~Leskovec, ``node2vec: Scalable feature learning for
  networks,'' in \emph{Proceedings of the 22nd ACM SIGKDD international
  conference on Knowledge discovery and data mining}, 2016, pp. 855--864.

\bibitem{zhang2018link}
M.~Zhang and Y.~Chen, ``Link prediction based on graph neural networks,''
  \emph{Advances in neural information processing systems}, vol.~31, 2018.

\bibitem{lee2019self}
J.~Lee, I.~Lee, and J.~Kang, ``Self-attention graph pooling,'' in
  \emph{International conference on machine learning}.\hskip 1em plus 0.5em
  minus 0.4em\relax PMLR, 2019, pp. 3734--3743.

\bibitem{gao2019graph}
H.~Gao and S.~Ji, ``Graph u-nets,'' in \emph{international conference on
  machine learning}.\hskip 1em plus 0.5em minus 0.4em\relax PMLR, 2019, pp.
  2083--2092.

\bibitem{ying2018hierarchical}
Z.~Ying, J.~You, C.~Morris, X.~Ren, W.~Hamilton, and J.~Leskovec,
  ``Hierarchical graph representation learning with differentiable pooling,''
  \emph{Advances in neural information processing systems}, vol.~31, 2018.

\bibitem{wu2022structural}
J.~Wu, X.~Chen, K.~Xu, and S.~Li, ``Structural entropy guided graph
  hierarchical pooling,'' in \emph{International conference on machine
  learning}.\hskip 1em plus 0.5em minus 0.4em\relax PMLR, 2022, pp.
  24\,017--24\,030.

\bibitem{bianchi2020spectral}
F.~M. Bianchi, D.~Grattarola, and C.~Alippi, ``Spectral clustering with graph
  neural networks for graph pooling,'' in \emph{International conference on
  machine learning}.\hskip 1em plus 0.5em minus 0.4em\relax PMLR, 2020, pp.
  874--883.

\bibitem{ju2024comprehensive}
W.~Ju, Z.~Fang, Y.~Gu, Z.~Liu, Q.~Long, Z.~Qiao, Y.~Qin, J.~Shen, F.~Sun,
  Z.~Xiao \emph{et~al.}, ``A comprehensive survey on deep graph representation
  learning,'' \emph{Neural Networks}, p. 106207, 2024.

\bibitem{vaswani2017attention}
A.~Vaswani, N.~Shazeer, N.~Parmar, J.~Uszkoreit, L.~Jones, A.~N. Gomez,
  {\L}.~Kaiser, and I.~Polosukhin, ``Attention is all you need,''
  \emph{Advances in neural information processing systems}, vol.~30, 2017.

\bibitem{liu2022abdpool}
Y.~Liu, L.~Cui, Y.~Wang, and L.~Bai, ``Abdpool: Attention-based differentiable
  pooling,'' in \emph{2022 26th International Conference on Pattern Recognition
  (ICPR)}.\hskip 1em plus 0.5em minus 0.4em\relax IEEE, 2022, pp. 3021--3026.

\bibitem{xu2018representation}
K.~Xu, C.~Li, Y.~Tian, T.~Sonobe, K.-i. Kawarabayashi, and S.~Jegelka,
  ``Representation learning on graphs with jumping knowledge networks,'' in
  \emph{International conference on machine learning}.\hskip 1em plus 0.5em
  minus 0.4em\relax PMLR, 2018, pp. 5453--5462.

\bibitem{abu2019mixhop}
S.~Abu-El-Haija, B.~Perozzi, A.~Kapoor, N.~Alipourfard, K.~Lerman,
  H.~Harutyunyan, G.~Ver~Steeg, and A.~Galstyan, ``Mixhop: Higher-order graph
  convolutional architectures via sparsified neighborhood mixing,'' in
  \emph{international conference on machine learning}.\hskip 1em plus 0.5em
  minus 0.4em\relax PMLR, 2019, pp. 21--29.

\bibitem{abu2020n}
S.~Abu-El-Haija, A.~Kapoor, B.~Perozzi, and J.~Lee, ``N-gcn: Multi-scale graph
  convolution for semi-supervised node classification,'' in \emph{uncertainty
  in artificial intelligence}.\hskip 1em plus 0.5em minus 0.4em\relax PMLR,
  2020, pp. 841--851.

\bibitem{bruna2013spectral}
J.~Bruna, W.~Zaremba, A.~Szlam, and Y.~LeCun, ``Spectral networks and locally
  connected networks on graphs,'' \emph{arXiv preprint arXiv:1312.6203}, 2013.

\bibitem{defferrard2016convolutional}
M.~Defferrard, X.~Bresson, and P.~Vandergheynst, ``Convolutional neural
  networks on graphs with fast localized spectral filtering,'' \emph{Advances
  in neural information processing systems}, vol.~29, 2016.

\bibitem{zhang2018gaan}
J.~Zhang, X.~Shi, J.~Xie, H.~Ma, I.~King, and D.-Y. Yeung, ``Gaan: Gated
  attention networks for learning on large and spatiotemporal graphs,''
  \emph{arXiv preprint arXiv:1803.07294}, 2018.

\bibitem{dobson2003distinguishing}
P.~D. Dobson and A.~J. Doig, ``Distinguishing enzyme structures from
  non-enzymes without alignments,'' \emph{Journal of molecular biology}, vol.
  330, no.~4, pp. 771--783, 2003.

\bibitem{borgwardt2005protein}
K.~M. Borgwardt, C.~S. Ong, S.~Sch{\"o}nauer, S.~Vishwanathan, A.~J. Smola, and
  H.-P. Kriegel, ``Protein function prediction via graph kernels,''
  \emph{Bioinformatics}, vol.~21, no. suppl\_1, pp. i47--i56, 2005.

\bibitem{wale2008comparison}
N.~Wale, I.~A. Watson, and G.~Karypis, ``Comparison of descriptor spaces for
  chemical compound retrieval and classification,'' \emph{Knowledge and
  Information Systems}, vol.~14, pp. 347--375, 2008.

\bibitem{orsini2015graph}
F.~Orsini, P.~Frasconi, and L.~De~Raedt, ``Graph invariant kernels,'' in
  \emph{Proceedings of the twenty-fourth international joint conference on
  artificial intelligence}, vol. 2015.\hskip 1em plus 0.5em minus 0.4em\relax
  IJCAI-INT JOINT CONF ARTIF INTELL, 2015, pp. 3756--3762.

\bibitem{yanardag2015deep}
P.~Yanardag and S.~Vishwanathan, ``Deep graph kernels,'' in \emph{Proceedings
  of the 21th ACM SIGKDD international conference on knowledge discovery and
  data mining}, 2015, pp. 1365--1374.

\bibitem{ranjan2020asap}
E.~Ranjan, S.~Sanyal, and P.~Talukdar, ``Asap: Adaptive structure aware pooling
  for learning hierarchical graph representations,'' in \emph{Proceedings of
  the AAAI conference on artificial intelligence}, vol.~34, no.~04, 2020, pp.
  5470--5477.

\bibitem{zhang2018end}
M.~Zhang, Z.~Cui, M.~Neumann, and Y.~Chen, ``An end-to-end deep learning
  architecture for graph classification,'' in \emph{Proceedings of the AAAI
  conference on artificial intelligence}, vol.~32, no.~1, 2018.

\bibitem{DBLP:conf/aaai/FengY0T22}
A.~Feng, C.~You, S.~Wang, and L.~Tassiulas, ``Kergnns: Interpretable graph
  neural networks with graph kernels,'' in \emph{Thirty-Sixth {AAAI} Conference
  on Artificial Intelligence, {AAAI} 2022, Thirty-Fourth Conference on
  Innovative Applications of Artificial Intelligence, {IAAI} 2022, The Twelveth
  Symposium on Educational Advances in Artificial Intelligence, {EAAI} 2022
  Virtual Event, February 22 - March 1, 2022}, 2022, pp. 6614--6622.

\bibitem{DBLP:conf/icml/ChenJM20}
D.~Chen, L.~Jacob, and J.~Mairal, ``Convolutional kernel networks for
  graph-structured data,'' in \emph{Proceedings of the 37th International
  Conference on Machine Learning, {ICML} 2020, 13-18 July 2020, Virtual Event},
  ser. Proceedings of Machine Learning Research, vol. 119, 2020, pp.
  1576--1586.

\bibitem{nair2010rectified}
V.~Nair and G.~E. Hinton, ``Rectified linear units improve restricted boltzmann
  machines,'' in \emph{Proceedings of the 27th international conference on
  machine learning (ICML-10)}, 2010, pp. 807--814.

\bibitem{szegedy2015going}
C.~Szegedy, W.~Liu, Y.~Jia, P.~Sermanet, S.~Reed, D.~Anguelov, D.~Erhan,
  V.~Vanhoucke, and A.~Rabinovich, ``Going deeper with convolutions,'' in
  \emph{Proceedings of the IEEE conference on computer vision and pattern
  recognition}, 2015, pp. 1--9.

\bibitem{kingma2014adam}
D.~P. Kingma and J.~Ba, ``Adam: A method for stochastic optimization,''
  \emph{arXiv preprint arXiv:1412.6980}, 2014.

\end{thebibliography}

\end{document}